\documentclass[sigconf]{acmart}

\AtBeginDocument{%
  \providecommand\BibTeX{{%
    \normalfont B\kern-0.5em{\scshape i\kern-0.25em b}\kern-0.8em\TeX}}}

\copyrightyear{2022}
\acmYear{2022}
\setcopyright{rightsretained}
\acmConference[GECCO '22 Companion]{Genetic and Evolutionary
Computation Conference Companion}{July 9--13, 2022}{Boston, MA, USA}
\acmBooktitle{Genetic and Evolutionary Computation Conference
Companion (GECCO '22 Companion), July 9--13, 2022, Boston, MA, USA}
\acmDOI{10.1145/3520304.3528922}
\acmISBN{978-1-4503-9268-6/22/07}

\usepackage{url}
\usepackage{graphicx}
\usepackage{subcaption}

\usepackage{algorithm}
\usepackage[noend]{algpseudocode}
\usepackage{setspace}
\let\Algorithm\algorithm
\renewcommand\algorithm[1][]{\Algorithm[#1]\setstretch{1.1}}

\algrenewcommand{\algorithmiccomment}[1]{\hskip3px$\#$ #1}
\algdef{SE}[SUBALG]{Indent}{EndIndent}{}{\algorithmicend\ }%
\algtext*{Indent}
\algtext*{EndIndent}
\usepackage{listings}
\begin{document}
\title{Binary and Multinomial Classification through\\Evolutionary Symbolic Regression}

\author{Moshe Sipper}
\email{sipper@bgu.ac.il}
\authornotemark[1]
\orcid{0000-0003-1811-472X}
\affiliation{%
  \institution{Department of Computer Science, Ben-Gurion University}
  \streetaddress{}
  \city{Beer Sheva 84105}
  \state{}
  \country{Israel}
  \postcode{84105}
}

\renewcommand{\shortauthors}{M. Sipper}

\begin{abstract}
We present three evolutionary symbolic regression-based classification algorithms for binary and multinomial datasets: GPLearnClf, CartesianClf, and ClaSyCo. 
Tested over 162 datasets and compared to three state-of-the-art machine learning algorithms---XGBoost, LightGBM, and a deep neural network---we find our algorithms to be competitive.
Further, we demonstrate how to find the best method for one's dataset automatically, through the use of a state-of-the-art hyperparameter optimizer. 
\end{abstract}

\begin{CCSXML}
<ccs2012>
   <concept>
       <concept_id>10010147.10010257.10010321</concept_id>
       <concept_desc>Computing methodologies~Machine learning algorithms</concept_desc>
       <concept_significance>500</concept_significance>
       </concept>
   <concept>
       <concept_id>10010147.10010257.10010293.10011809.10011813</concept_id>
       <concept_desc>Computing methodologies~Genetic programming</concept_desc>
       <concept_significance>500</concept_significance>
       </concept>
   <concept>
       <concept_id>10010147.10010257.10010293.10010294</concept_id>
       <concept_desc>Computing methodologies~Neural networks</concept_desc>
       <concept_significance>500</concept_significance>
       </concept>
 </ccs2012>
\end{CCSXML}

\ccsdesc[500]{Computing methodologies~Machine learning algorithms}
\ccsdesc[500]{Computing methodologies~Genetic programming}
\ccsdesc[500]{Computing methodologies~Neural networks}

\keywords{classification, genetic programming, symbolic regression}

\maketitle

\section{Introduction}
\label{sec:intro}
Classification is an important subfield of supervised learning. As such, many powerful algorithms have been designed over the years to tackle both binary datasets as well as multinomial, or multiclass ones.

Symbolic regression (SR) is a family of algorithms that aims to find regressors of arbitrary complexity. 
Herein, we show that evolutionary SR-based \textit{regressors} can be successfully converted into performant \textit{classifiers}.

% In the next section we provide background and survey some literature on evolutionary classification through symbolic regression.
% Section~\ref{sec:algs} presents three evolutionary SR-based algorithms of our own design:
% GPLearnClf, CartesianClf, and ClaSyCo.
% Section~\ref{sec:setup} delineates the experimental set-up used to test our new algorithms, followed by results and discussion in Section~\ref{sec:results}.
% Finally, we offer concluding remarks in Section~\ref{sec:conc}.

\section{Background and Previous Work}
\label{sec:prev}
Binary classification, wherein an input vector $\mathbf{X}$ is to be classified into one of two classes, $y \in \{0,1\}$, has received much attention in the literature. While some methods, e.g., decision trees, lend themselves naturally to an extension beyond binary to multinomial classification, regressors are perhaps somewhat less-natural multinomial classifiers. 

The well-known logistic regression algorithm passes the output of a linear regressor through a sigmoid function, $f(z) = \frac{1}{1+e^{-z}}$, and uses a cross-entropy loss function during training: 
$-(y\log(\hat{y})+(1-y)\log(1-\hat{y}))$, where $y \in \{0,1\}$ is the binary target label and $\hat{y}\in[0,1]$ is the model's predicted output (which can be treated as the probability assigned by the model of the output's belonging to class 1). A lower loss value is better, with zero being the best.

A common approach to extend binary to multinomial classification, where there are $C$ classes, $C>2$, is to use the \textit{one-vs-rest} (or \textit{one-vs-all}) method, in which we train $C$ binary classifiers, where the data from class $c$, $c \in \{1,\ldots ,C\}$, is treated as positive (or class 1), and the data from all other classes is treated as negative (or class 0).
This approach requires that each model predict a class membership probability or a probability-like score. The $\mathit{argmax}$ of these scores (class index with the largest score) is then used to predict a class.

In some cases, e.g., the well-known logistic regression algorithm, true multinomial classification is possible. %\cite{engel1988polytomous}. 
Multinomial logistic regression (also known as softmax regression) is a generalization of binary logistic regression to the case of multiple classes. In this case the target label $y$ can take on $C$ different values, with $C>2$. Given an input vector $\mathbf{X}$, we want our model to output a $C$-dimensional vector whose elements sum to 1, thus providing us with $C$ probability estimates. Prediction can then be done by using the $\mathit{argmax}$ function over all $C$ values, i.e., by outputting the class with maximum probability value.

The logistic-regression model, or hypothesis, computes $P(\hat{y}_c=c) = \sigma(o\mathbf{(X)})$, $\mathit{for}$ $c = 1,\ldots ,C$, where $o$ is the output of the underlying linear regressor, and  $\sigma: \mathbb{R}^C \to [0,1]^C$ is the softmax function:
$
\sigma(\mathbf{z})_i = \frac{e^{z_i}}{\sum_{j=1}^C e^{z_j}}, \;
i = 1,\ldots ,C; \;
 \;
\mathbf{z}=(z_1,\ldots,z_C) \in \mathbb{R}^C.
$
This is in fact similar to the operation of a neural network with a softmax output layer. The cost function used by softmax regression for model learning is cross-entropy, which maximizes the probability of the output vectors to the one-hot-encoded target $y$ values.
%As an aside, the highly popular Scikit-learn Python machine learning software library offers both options for the function LogisticRegression \cite{scikit-learn}.
%``\ldots the optimization problem is decomposed in a one-vs-rest fashion so separate binary classifiers are trained for all classes... Setting multi\_class to `multinomial' with these solvers learns a true multinomial logistic regression.'' \cite{sklearn-website}

Tree-based genetic programming (GP) evolves computational program trees that can be evaluated in a recursive manner. %A sample tree is shown in Figure~\ref{fig:gptree}---it is equivalent to the expression $x^2 + x + 1$.
Given a set of functions (internal tree nodes) and terminals (tree leaves), driven by a quality (fitness) function, and using stochastic, tree-modifying operators, GP is able to produce successively better models in an iterative manner. GP thus evolves regression models, an approach known as symbolic regression (SR). % \cite{koza1992genetic,schmidt2009distilling,korns2016highly}. 

% \begin{figure}
% \centering
% \includegraphics[scale=0.45]{gptree.png}
% \caption{A GP tree, which is equivalent to the expression $x^2 + x + 1$. The tree consists of inner nodes taken from a pre-defined function set and leaves comprising problem features and constants. The tree's value is computed recursively in a depth-first manner, and the final output is that computed by the root node.}
% \label{fig:gptree}
% \end{figure}

If one specifies a threshold then the output of an evolved tree can be converted into a binary classification value, depending on whether the output is smaller or larger than the threshold. Setting the threshold judiciously may not always be straightforward.

\cite{espejo2009survey} presented an excellent survey on the application of GP to classification, for both the binary and the multinomial cases.
They reviewed several multinomial GP-based classification algorithms that essentially followed the one-vs-rest scheme. Another scheme involved the use of a single classification function, along with $C-1$ threshold values, which determined $C$ intervals, each one assigned to a different class.

\cite{Korns2019} recently proposed an enhanced SR classification approach, LDA$^{++}$, comparing it to four SR-based approaches and five non-SR approaches. 
The latter were: multi-layer perceptron, decision tree, random forest, tree ensemble, and gradient boosting. The five SR classifiers represent, in fact, what might be considered the state-of-the-art in evolutionary SR-based classification:
\begin{enumerate}
    \item Evolve $C$ separate GP functions, one per each of $C$ classes. 
    The output then equals:
    $\hat{y}=\mathit{argmax}(\mathit{gp}_1, \mathit{gp}_2,\ldots, \mathit{gp}_C)$.
    
    \item Multilayer Discriminant Classification (MDC) enhances the previous approach by calculating:
    \begin{align*}
    \hat{y}=\mathit{argmax}(w_{10}+w_{11}\mathit{gp}_1, \ldots, w_{C0}+w_{C1}\mathit{gp}_C).
    \end{align*}
    MDC optimizes the coefficient weights, $\{w_{10}, w_{11}, ..., w_{C1}\}$.
    % $\{w_{10}, w_{11}, \ldots, w_{C0}, w_{C1}\}$.
   
    \item The M$_2$GP algorithm, which basically generates a $d$-dimens-ional GP tree instead of a 1-dimensional tree \cite{m2gp}. %The GP tree's root node is modified to have an arity of $d$ ($d \geq 1$). The root node cannot be modified through genetic operators. Essentially, an input is mapped into a $d$-dimensional solution space. A mapped sample is predicted to belong to class $c$ if it has the minimum Mahalanobis distance measured against the centroid of the $c$th $C$-clustered mapped data.
    
    \item Evolutionary Linear Discriminant Analysis (LDA). LDA is a statistical method used to find linear combinations of features separating two or more classes. \cite{korns2017evolutionary} combined LDA and GP symbolic classification for financial multiclass classification problems.
    
    \item GP-assisted LDA was enhanced with a modified version of Platt’s Sequential Minimal Optimization algorithm % \cite{platt1998sequential} 
    and with swarm-optimization techniques. % \cite{karaboga2009survey}. 
    Adding a user-defined typing system and deep-learning feature selection resulted in the LDA$^{++}$ algorithm. 
    
    This algorithm was compared with the other SR and non-SR algorithms above over 10 artificial classification problems \cite{korns2018genetic}. Of these 10, LDA$^{++}$ achieved the top result for 6 datasets, gradient boosting won 2 datasets, multi-layer perceptron and MDC -- each 1 dataset. Some of the wins were by a small margin and the statistical significance is unclear. A further real-world banking dataset was examined, and LDA$^{++}$ came second to gradient boosting by a margin of 2\%. Thus, while LDA$^{++}$ showed promise, ``a great deal more work has to be done'' \cite{Korns2019}. 
    
\end{enumerate}

The above evolutionary SR-based algorithms are not publicly available through online repositories and their performance has been tested on a small number of datasets with mixed results.
This led us to focus our comparison in Section~\ref{sec:results} on three top, state-of-the art, machine-learning classifiers: 
XGBoost (Extreme Gradient Boosting) \cite{Chen:2016},
LightGBM (Light Gradient Boosting Machine) \cite{ke2017lightgbm}, and a
Deep Neural Network (DNN), with with 10 hidden layers of size 16 nodes each.

% \begin{enumerate}
% \item XGBoost \cite{Chen:2016}: Extreme Gradient Boosting. This performant algorithm has been used successfully on a multitude of hard problems, including chronic kidney disease diagnosis \cite{ogunleye2019xgboost}, trade in the financial markets \cite{nobre2019combining}, and network intrusion detection \cite{devan2020efficient}.
    
% \item LightGBM \cite{ke2017lightgbm}: Light Gradient Boosting Machine. Another state-of-the-art, potent algorithm, used to address many hard tasks, including cryptocurrency price trend forecasting \cite{sun2020novel}, prediction of chemical toxicity \cite{zhang2019lightgbm}, and prediction of blood-brain-barrier penetration \cite{shaker2021lightbbb}.

% \item Deep Neural Networks (DNNs) \cite{lecun2015deep}, with with 10 hidden layers of size 16 nodes each. DNNs have been applied successfully to a plethora of complex problems, including multiple sclerosis lesion detection and segmentation \cite{nair2020exploring}, wireless  resource allocation with application to vehicular networks \cite{liang2019deep}, and the study of multimodal brain development \cite{hu2019deep}.
% \end{enumerate}

\section{Algorithms}
\label{sec:algs}
We devised and tested three evolutionary SR-based classifiers: GPLearnClf, CartesianClf, and ClaSyCo. The first two are based on the one-vs-rest approach, while the last one is inherently multinomial. 

\textit{GPLearnClf} is based on the GPLearn package \cite{stephens2019gplearn}, which implements tree-based GP symbolic regression, is relatively fast, and---importantly---interfaces seamlessly with Scikit-learn \cite{scikit-learn}.
GPLearnClf evolves $C$ separate populations independently, each fitted to a specific class by considering as target values the respective column vector (of $C$ column vectors) of the one-hot-encoded target vector $y$. The fitness function is based on log loss (aka binary cross-entropy). Prediction is carried out by outputting the argmax of the set of best evolved individuals (one from each population). The hyperparameters to tune were population size, \textit{n\_pop}, and generation count, \textit{n\_gens}. We will discuss hyperparameters in Section~\ref{sec:setup} (for our SR-based classifiers, the same hyperparameter values were used by all $C$ populations).

\textit{CartesianClf} is based on Cartesian GP, which grew from a method of evolving digital circuits \cite{miller2011cartesian}. It is called `Cartesian' because it represents a program using a two-dimensional grid of nodes. 
The CGP package we used \cite{Ohjeah} evolves the population in a $(1 + \lambda)$-manner, i.e., in each generation it creates $\lambda$ offspring (we used the default $\lambda =4$) and compares their fitness to the parent individual. The fittest individual carries over to the next generation; in case of a draw, the offspring is preferred over the parent. Tournament selection is used (tournament size $=|population|$), single-point mutation, and no crossover. 
%For more details, see \cite{Ohjeah,miller2006redundancy}.

We implemented CartesianClf similarly to GPLearnClf in a one-vs-rest manner, with $C$ separate populations evolving independently, using binary cross-entropy as fitness.
The hyperparameters to tune were number of rows, \textit{n\_rows}, number of columns, \textit{n\_columns}, and maximum number of generations, \textit{maxiter}.

\textit{ClaSyCo} (\textbf{Cla}ssification through \textbf{Sy}mbolic Regression and\linebreak \textbf{Co}evolution) also employs $C$ populations of trees;
however, these are not evolved independently as with the one-vs-rest method (as done with GPLearnClf and CartesianClf)---but in tandem through \textit{cooperative coevolution}.

A cooperative coevolutionary algorithm involves a number of evolving populations, which come together to obtain problem solutions. The fitness of an individual in a particular population depends on its ability to collaborate with individuals from the other populations \cite{Pena2001,sipper2019omnirep}.

Specifically, in our case,
an individual SR tree $i$ in population $c$, $\mathit{gp}^c_i$, $i \in \mathit{\{1,\ldots,n\_pop\}}$, $c \in \{1,\ldots,C\}$, is assigned fitness through the following steps (we describe this per single dataset sample, although in practice fitness computation is vectorized by Python):
\begin{enumerate}
    \item Individual $\mathit{gp}^c_i$ computes an output $\hat{y}^c_i$ for the sample under consideration.
    
    \item Obtain the best-fitness classifier of the previous generation, $\mathit{gp}^{c'}_{\mathit{best}}$, for each population $c'$, $c' \in \{1,\ldots,C\}$, $c' \neq c$
    (these are called ``representatives'' or ``cooperators'' \cite{Pena2001}).
    
    \item Each $\mathit{gp}^{c'}_{\mathit{best}}$ computes an output $\hat{y}^{c'}_{\mathit{best}}$ for the sample under consideration.
    
    \item We now have $C$ output values, $\hat{y}^{1}_{\mathit{best}}, ... , \hat{y}^c_i , ... , \hat{y}^{C}_{\mathit{best}}$. 
    
    \item Compute $\sigma(\hat{y}^{1}_{\mathit{best}}, ... , \hat{y}^c_i , ... , \hat{y}^{C}_{\mathit{best}})$, where $\sigma$ is the softmax function.
    
    \item Assign a fitness score to $\mathit{gp}^c_i$ using the cross-entropy loss function. (NB: only individual $\mathit{gp}^c_i$ is assigned fitness---all other $C-1$ individuals are representatives.)
\end{enumerate}

Note that an individual in a single population---charged with classifying instances as to whether they belong to the respective (single) class or not---cannot obtain a fitness value without cooperating with all other $C-1$ populations; this is different than the one-vs-rest approach, where the populations evolve entirely independently (including fitness computation).
Essentially, we are treating the outputs of $C$ individuals analogously to a $C$-sized
softmax output layer of a neural network.

Aside from fitness computation, done in a cooperative manner, all other evolutionary operations (selection, crossover, mutation) are done per population exactly as with standard single-population evolution. The hyperparameters to tune for ClaSyCo were population size, \textit{n\_pop}, and generation count, \textit{n\_gens}.

GPLearnClf, CartesianClf, and ClaSyCo used standard mathematical operations in the function set: add, sub, mul, div, sqrt, log, abs, neg, min, max (div, sqrt, and log were protected versions of the underlying functions, to prevent illegal operations). The terminals were the problem features, which depended on the particular dataset being used.

\section{Experimental Setup}
\label{sec:setup}

For our experiments we used the popular Scikit-learn Python package \cite{scikit-learn} %,sklearn-website} 
due to its superb ability to handle much of the routine desiderata of machine learning coding and experimentation. We compared our three proposed classifiers to the three state-of-the-art ones discussed in Section~\ref{sec:prev}.
Thus, our experiment comprised a comparison of six classifiers: 
GPLearnClf, CartesianClf, ClaSyCo, XGBoost, LightGBM, and DNN.

We used Optuna, a state-of-the-art automatic hyperparameter optimization software framework \cite{akiba2019optuna}. Optuna offers a define-by-run style user API where one can dynamically construct the search space, and an efficient sampling algorithm and pruning algorithm. Moreover, our experience has shown it to be fairly easy to set up.

Optuna formulates the hyperparameter optimization problem as a process of minimizing or maximizing an objective function that takes a set of hyperparameters as an input and returns its (validation) score. We used the default Tree-structured Parzen Estimator (TPE) sampling algorithm. Optuna also provides pruning: automatic early stopping of unpromising trials \cite{akiba2019optuna}.

Optuna was tasked with performing the hyperparameter search. %through the space spanned by the values in Table~\ref{tab:params}. 
Further, we added the six classifiers themselves into Optuna's mix---as part of the hyperparameter search space. To wit, Optuna was charged with finding \textit{the best classifier} along with its hyperparameters. 
A sample output was of the form:
\begin{lstlisting}
    {'classifier': 'ClaSyCo', 'ClaSyCo_n_pop': 27, 'ClaSyCo_n_gens': 135}
\end{lstlisting}
This means that Optuna found ClaSyCo to be the best classifier, along with hyperparameters n\_pop=27 and n\_gens=135.
The hyperparameter value ranges (for numerical values) or sets (for categorical values) used by Optuna can be found as part of our code, which is available in its entirety at  \url{github.com/moshesipper}.

% \begin{table}
% \caption{Hyperparameter value ranges (for numerical values) or sets (for categorical values) used by Optuna.
%          Med/Mod column shows the medians (for numerical values) or modes (for categorical values) of Optuna-selected values in the first experiment of Section~\ref{sec:results}. }
% \label{tab:params}         
% \centering
% % \vspace{5pt}
% \resizebox{0.48\textwidth}{!}{%
% \begin{tabular}{r|c|c|l}
% \textbf{Algorithm} & \textbf{Hyperparameter} & \textbf{Range/Set} & \textbf{Med/Mod} \\ \hline 

% GPLearnClf & n\_pop & [20, 200] & 125 \\
%           & n\_gens & [20, 200] & 112\\ \hline

% CartesianClf & n\_rows & [1, 10] & 5 \\
%              & n\_columns & [1, 10] & 5 \\ 
%              & maxiter & [10, 1000] & 83 \\ \hline

% ClaSyCo & n\_pop & [20, 200] & 124 \\
%         & n\_gens & [20, 200] & 136 \\ \hline
        
% XGBoost & n\_estimators & [10, 1000] & 158 \\ 
%         & learning\_rate & [0.01, 0.2] & 0.12 \\
%         & gamma & [0, 0.4] & 0.2 \\
%         & max\_depth & [4, 6] & 5 \\
%         & subsample & [0.5, 1] & 0.77 \\ \hline
        
% LightGBM & n\_estimators & [10, 1000] & 207 \\ 
%          & learning\_rate & [0.01, 0.2] & 0.12 \\
%          & max\_depth & [4, 6] & 5 \\
%          & bagging\_fraction & [0.5, 0.95] & 0.73 \\ 
%          & bagging\_freq & [1, 10] & 5 \\ \hline

% DNN & activation & \{`identity', `logistic', `tanh', `relu'\} & `tanh' \\
%     & solver & \{`lbfgs', `sgd', `adam'\} & `adam' \\
%     & learning\_rate & \{`constant', `invscaling', `adaptive'\} & `constant' \\ \hline

% \end{tabular}
% }
% \end{table}

The pseudo-code of the experimental setup is given in Algorithm~\ref{alg:setup}. 
Each single-dataset replicate run begins with an 80\%-20\% random train-test split.
We then fit Scikit-learn's StandardScaler to the training set and apply the fitted scaler to the test set. This ensures that features have zero mean and unit variance (often helpful for non-tree-based algorithms).

\begin{algorithm}
\small
\caption{Experimental setup (per dataset)}\label{alg:setup}
\begin{algorithmic}[1]
\Statex
\Require
\Indent
\Statex \textit{dataset} $\gets$ dataset to be used
\Statex \textit{classifiers} $\gets$ \footnotesize{\{GPLearnClf, CartesianClf, ClaSyCo, XGBoost, LightGBM, DNN\} }
\EndIndent

\Ensure
\Indent
\Statex Best algorithm (per each replicate run)
\EndIndent

\For{\textit{replicate} $\gets$ 1 to \textit{20}} 
  \State Randomly split \textit{dataset} into 80\% \textit{training set} and 20\% \textit{test set}
  \State Fit StandardScaler to \textit{training set} and apply fitted scaler to \textit{test set}
  \State Run Optuna for 100 trials %with hyperparameter space spanned by  \textit{classifiers} along with their hyperparameters (Table~\ref{tab:params})
  \State Record \textit{best algorithm} returned by Optuna
\EndFor
\end{algorithmic}
\normalsize
\end{algorithm} 

A single Optuna trial represents a process of evaluating an objective function, where Optuna provides hyperparameter suggestions when requested.
In our case, in each trial Optuna was asked to suggest a classifier along with hyperparameters for it. The classifier was then fit to the training data; 
the fitted classifier's balanced accuracy score over the left-out test data
was returned to Optuna  as the trial's score. 

Note that we did not use a third left-out data fold, as our study focused on having Optuna find the best-performing classifier on unseen (test) data. We were interested in ascertaining, through Optuna, which classifier emerged as winner per replicate run, where winning was over the test set. Thus, there was no need for further data splitting within the context of our investigation. 

Our use of Optuna represents a novel assessment and comparison method.
Further, it serves to demonstrate a practical modus operandi when one wishes to obtain the best algorithm for a particular real-life dataset.

\section{Results and Discussion}
\label{sec:results}

%We ran our experiments on our cluster of Intel\textsuperscript{\textregistered} Xeon\textsuperscript{\textregistered} E5-2650L servers. 
We used PMLB \cite{Olson2017PMLB}, 
which offers 162 curated classification datasets---90 binary datasets and 72 multinomial datasets---with number of samples between 32--1025010, number of features between 2--1000, and number of classes between 2--26.
% While our main aim herein was to do well on multinomial datasets ($C>2$), we experimented with the binary ones as well.
Results are presented separately for binary and multinomial datasets.

\newcommand\binruns{1653 }
\newcommand\multruns{869 }

Each dataset was run 20 times for up to 48 hours---whichever came first. 
% As noted above, each run used a different random train-test split of 80\%-20\%, respectively. 
We set the number of trials used by Optuna to 100 per dataset per run.
% Each trial consisted of training an Optuna-suggested model on the training set and then returning (to Optuna) the balanced accuracy score over the test set as the trial's score.
When the 48 hours were up we retained the datasets for which at least half the runs (i.e., 10) finished. 
This resulted in a total of 132 datasets---84 binary datasets and
48 multinomial ones---with \binruns independent runs for the binary datasets 
and \multruns independent runs for the multinomial datasets.

Table~\ref{tab:ranking} presents the percentage of times each algorithm was chosen as best by Optuna.
Unsurprisingly, XGBoost, LightGBM, and DNN performed very well.
Our 3 SR-based classifiers were selected as best classifier for a total of 
32.6\% for the binary case and 13.8\% for the multinomial case.
While this latter is a drop from the former, both values are still very respectable, given the competing state-of-the-art algorithms.  

\begin{table}
\caption{Percentage of times each algorithm was chosen by Optuna as best (win), 
        of the \binruns independent  runs for  the  binary  datasets,  
        and \multruns independent runs for the multinomial datasets.}
\label{tab:ranking}
\centering
\scriptsize
\resizebox{0.49\textwidth}{!}{%
\begin{tabular}{cc}
{\small\bf Binary} & {\small\bf Multinomial} \\[2pt]
     
\begin{tabular}{r|l} 
Algorithm & Wins \\ \hline
XGBoost & 28.13\% \\  
LightGBM & 24.5\% \\
DNN & 14.76\% \\
ClaSyCo & 13.85\% \\
CartesianClf & 9.44\% \\
GPLearnClf & 9.32\% \\
\end{tabular}

&

\begin{tabular}{r|l} 
Algorithm & Wins \\ \hline
LightGBM & 38.55\% \\  
XGBoost & 32.8\% \\
DNN & 14.84\% \\
ClaSyCo & 6.1\% \\
GPLearnClf & 4.14\% \\
CartesianClf & 3.57\% \\
\end{tabular}

\normalsize
\end{tabular}
}
\end{table}

%Table~\ref{tab:params} shows the medians (for numerical values) and modes (for categorical values) of the hyperparameter values selected by Optuna (across all independent runs, both binary and multinomial).

Note that within our Optuna-based scenario we do not perform a ranking of all algorithms, for which we can asses statistical significance of, say, the different positions. 
Rather, we task Optuna with finding the best classifier, and simply tally up the number of times each algorithm is chosen as Optuna's answer. 

Is there an underlying pattern that would help us select the best algorithm given basic dataset attributes: no. samples, no. features, no. classes?
To answer this question we created a dataset from our results, where each row comprised 3 attributes---no. samples, no. features, no. classes---and the target value was the classifier chosen as best by Optuna. 

We then ran 12 machine learning algorithms---the 6 used herein plus 6 others: gradient boosting, ridge classifier, logistic regression, random forest, AdaBoost and decision tree---but were unable to obtain a good prediction model that would predict the most appropriate classifier to use: the best balanced test-set accuracy over 50 independent runs was only 0.35. 
This might well be improved in future work if we take into account more dataset attributes.
The good news for now, though, is that our Optuna-based methodology is able to automatically discover the top algorithm for a given dataset (along with its hyperparameters).

How do the three evolutionary SR algorithms compare amongst themselves?
To answer this, we ran the entire experiment (162 datasets, 20 runs per dataset, 48-hour limit) with only GPLearnClf, CartesianClf, and ClaSyCo. When  the 48  hours  were  up  we  retained  the  datasets  for which  at  least  half the  runs finished, resulting in a total of 
61 binary datasets (1159 independent runs) and 18 multinomial datasets (282 independent runs).
For the binary datasets results were: 
ClaSyCo -- 47.11\% wins, 
GPLearnClf -- 26.83\%,
CartesianClf -- 26.06\%;
for the multinomial datasets:
ClaSyCo -- 45.39\%,  
GPLearnClf -- 31.21\%, 
CartesianClf -- 23.4\%.

% Table~\ref{tab:SR} shows the win percentages; 
% %In relative terms they are similar to those of the full experiment (Table~\ref{tab:ranking}).

% \begin{table}
% \caption{Win percentages when running only GPLearnClf, CartesianClf, and ClaSyCo.}
% \label{tab:SR}
% \centering
% \scriptsize
% \resizebox{0.49\textwidth}{!}{%
% \begin{tabular}{cc}
% {\small\bf Binary} & {\small\bf Multinomial} \\[2pt]
     
% \begin{tabular}{r|l} 
% Algorithm & Wins \\ \hline
% ClaSyCo & 47.11\% \\  
% GPLearnClf & 26.83\% \\
% CartesianClf & 26.06\% \\
% \end{tabular}

% &

% \begin{tabular}{r|l}  
% Algorithm & Wins \\ \hline
% ClaSyCo & 45.39\% \\ 
% GPLearnClf & 31.21\% \\
% CartesianClf & 23.4\% \\
% \end{tabular}

% \normalsize
% \end{tabular}
% }
% \end{table}

\section{Concluding Remarks}
\label{sec:conc}
We presented three evolutionary symbolic regression-based classification algorithms and showed that they perform well over binary and multinomial datasets, able to surpass state-of-the-art algorithms for a significant number of dataset runs.
We demonstrated how to find the best method for one's dataset automatically, through the use of a state-of-the-art hyperparameter optimizer. 

We noted that our evolutionary SR algorithms emerged as best for over one third of the binary datasets and about half that number for the multinomial case.
This means that often enough they may prove useful, by surpassing top ML methods, and we think they would be a worthy addition to the ML toolkit of classification algorithms.

We believe this is a promising approach with immediate applicability.
Further, there are a number of interesting paths to be explored:
\begin{itemize}
    \item Allow more resources---both Optuna number of trials as well as time limit---to see whether significant improvement can be attained.   
    \item Consider other symbolic regression algorithms.
    \item Develop an algorithm recommender that takes into account more than just no. samples, no. features, and no. classes, as done above. We might also consider some key aspects of datasets not easily observable, i.e., the nature of the underlying patterns of association (univariate, additive, interactions, heterogeneity, number of true predictive features, etc.).
    %As noted in the previous section, adding more dataset attributes may well improve the recommender's performance.
\end{itemize}

\section*{Acknowledgements}
I thank Snir Vitrack Tamam and the anonymous reviewers for helpful comments.

\bibliography{refs}


\begin{thebibliography}{15}

%%% the Web address.

\ifx \showCODEN    \undefined \def \showCODEN     #1{\unskip}     \fi
\ifx \showDOI      \undefined \def \showDOI       #1{#1}\fi
\ifx \showISBNx    \undefined \def \showISBNx     #1{\unskip}     \fi
\ifx \showISBNxiii \undefined \def \showISBNxiii  #1{\unskip}     \fi
\ifx \showISSN     \undefined \def \showISSN      #1{\unskip}     \fi
\ifx \showLCCN     \undefined \def \showLCCN      #1{\unskip}     \fi
\ifx \shownote     \undefined \def \shownote      #1{#1}          \fi
\ifx \showarticletitle \undefined \def \showarticletitle #1{#1}   \fi
\ifx \showURL      \undefined \def \showURL       {\relax}        \fi
% invisible to TeX
\providecommand\bibfield[2]{#2}
\providecommand\bibinfo[2]{#2}
\providecommand\natexlab[1]{#1}
\providecommand\showeprint[2][]{arXiv:#2}

\bibitem[\protect\citeauthoryear{Akiba, Sano, Yanase, Ohta, and Koyama}{Akiba
  et~al\mbox{.}}{2019}]%
        {akiba2019optuna}
\bibfield{author}{\bibinfo{person}{Takuya Akiba}, \bibinfo{person}{Shotaro
  Sano}, \bibinfo{person}{Toshihiko Yanase}, \bibinfo{person}{Takeru Ohta},
  {and} \bibinfo{person}{Masanori Koyama}.} \bibinfo{year}{2019}\natexlab{}.
\newblock \showarticletitle{Optuna: A next-generation hyperparameter
  optimization framework}. In \bibinfo{booktitle}{\emph{Proceedings of the 25th
  ACM SIGKDD international conference on knowledge discovery \& data mining}}.
  \bibinfo{pages}{2623--2631}.
\newblock


\bibitem[\protect\citeauthoryear{Chen and Guestrin}{Chen and Guestrin}{2016}]%
        {Chen:2016}
\bibfield{author}{\bibinfo{person}{Tianqi Chen} {and} \bibinfo{person}{Carlos
  Guestrin}.} \bibinfo{year}{2016}\natexlab{}.
\newblock \showarticletitle{{XGBoost}: A Scalable Tree Boosting System}. In
  \bibinfo{booktitle}{\emph{Proceedings of the 22nd ACM SIGKDD International
  Conference on Knowledge Discovery and Data Mining}} (San Francisco,
  California, USA) \emph{(\bibinfo{series}{KDD '16})}. \bibinfo{address}{New
  York, NY, USA}, \bibinfo{pages}{785--794}.
\newblock


\bibitem[\protect\citeauthoryear{Espejo, Ventura, and Herrera}{Espejo
  et~al\mbox{.}}{2009}]%
        {espejo2009survey}
\bibfield{author}{\bibinfo{person}{Pedro~G Espejo},
  \bibinfo{person}{Sebasti{\'a}n Ventura}, {and} \bibinfo{person}{Francisco
  Herrera}.} \bibinfo{year}{2009}\natexlab{}.
\newblock \showarticletitle{A survey on the application of genetic programming
  to classification}.
\newblock \bibinfo{journal}{\emph{IEEE Transactions on Systems, Man, and
  Cybernetics, Part C (Applications and Reviews)}} \bibinfo{volume}{40},
  \bibinfo{number}{2} (\bibinfo{year}{2009}), \bibinfo{pages}{121--144}.
\newblock


\bibitem[\protect\citeauthoryear{GPLearn}{GPLearn}{2021}]%
        {stephens2019gplearn}
GPLearn \bibinfo{year}{2021}\natexlab{}.
\newblock \bibinfo{title}{{GPLearn}}.
\newblock \bibinfo{howpublished}{\url{https://gplearn.readthedocs.io/}}.
\newblock
\newblock
\shownote{Accessed: 2021-4-30.}


\bibitem[\protect\citeauthoryear{Ingalalli, Silva, Castelli, and
  Vanneschi}{Ingalalli et~al\mbox{.}}{2014}]%
        {m2gp}
\bibfield{author}{\bibinfo{person}{Vijay Ingalalli}, \bibinfo{person}{Sara
  Silva}, \bibinfo{person}{Mauro Castelli}, {and} \bibinfo{person}{Leonardo
  Vanneschi}.} \bibinfo{year}{2014}\natexlab{}.
\newblock \showarticletitle{A Multi-dimensional Genetic Programming Approach
  for Multi-class Classification Problems}. In
  \bibinfo{booktitle}{\emph{Genetic Programming}},
  \bibfield{editor}{\bibinfo{person}{Miguel Nicolau},
  \bibinfo{person}{Krzysztof Krawiec}, \bibinfo{person}{Malcolm~I. Heywood},
  \bibinfo{person}{Mauro Castelli}, \bibinfo{person}{Pablo
  Garc{\'i}a-S{\'a}nchez}, \bibinfo{person}{Juan~J. Merelo},
  \bibinfo{person}{Victor~M. Rivas~Santos}, {and} \bibinfo{person}{Kevin Sim}}
  (Eds.). \bibinfo{publisher}{Springer Berlin Heidelberg},
  \bibinfo{address}{Berlin, Heidelberg}, \bibinfo{pages}{48--60}.
\newblock


\bibitem[\protect\citeauthoryear{Ke, Meng, Finley, Wang, Chen, Ma, Ye, and
  Liu}{Ke et~al\mbox{.}}{2017}]%
        {ke2017lightgbm}
\bibfield{author}{\bibinfo{person}{Guolin Ke}, \bibinfo{person}{Qi Meng},
  \bibinfo{person}{Thomas Finley}, \bibinfo{person}{Taifeng Wang},
  \bibinfo{person}{Wei Chen}, \bibinfo{person}{Weidong Ma},
  \bibinfo{person}{Qiwei Ye}, {and} \bibinfo{person}{Tie-Yan Liu}.}
  \bibinfo{year}{2017}\natexlab{}.
\newblock \showarticletitle{{LightGBM}: A highly efficient gradient boosting
  decision tree}.
\newblock \bibinfo{journal}{\emph{Advances in Neural Information Processing
  Systems}}  \bibinfo{volume}{30} (\bibinfo{year}{2017}),
  \bibinfo{pages}{3146--3154}.
\newblock


\bibitem[\protect\citeauthoryear{Korns}{Korns}{2017}]%
        {korns2017evolutionary}
\bibfield{author}{\bibinfo{person}{Michael~F Korns}.}
  \bibinfo{year}{2017}\natexlab{}.
\newblock \showarticletitle{Evolutionary linear discriminant analysis for
  multiclass classification problems}. In \bibinfo{booktitle}{\emph{Proceedings
  of the Genetic and Evolutionary Computation Conference Companion}}.
  \bibinfo{pages}{233--234}.
\newblock


\bibitem[\protect\citeauthoryear{Korns}{Korns}{2018}]%
        {korns2018genetic}
\bibfield{author}{\bibinfo{person}{Michael~F Korns}.}
  \bibinfo{year}{2018}\natexlab{}.
\newblock \showarticletitle{Genetic programming symbolic classification: A
  study}.
\newblock In \bibinfo{booktitle}{\emph{Genetic Programming Theory and Practice
  XV}}. \bibinfo{publisher}{Springer}, \bibinfo{pages}{39--54}.
\newblock


\bibitem[\protect\citeauthoryear{Korns and May}{Korns and May}{2019}]%
        {Korns2019}
\bibfield{author}{\bibinfo{person}{Michael~F. Korns} {and} \bibinfo{person}{Tim
  May}.} \bibinfo{year}{2019}\natexlab{}.
\newblock \showarticletitle{Strong Typing, Swarm Enhancement, and Deep Learning
  Feature Selection in the Pursuit of Symbolic Regression-Classification}. In
  \bibinfo{booktitle}{\emph{Genetic Programming Theory and Practice XVI}},
  \bibfield{editor}{\bibinfo{person}{Wolfgang Banzhaf}, \bibinfo{person}{Lee
  Spector}, {and} \bibinfo{person}{Leigh Sheneman}} (Eds.).
  \bibinfo{publisher}{Springer International Publishing},
  \bibinfo{address}{Cham}, \bibinfo{pages}{59--84}.
\newblock


\bibitem[\protect\citeauthoryear{Miller}{Miller}{2011}]%
        {miller2011cartesian}
\bibfield{author}{\bibinfo{person}{Julian~F Miller}.}
  \bibinfo{year}{2011}\natexlab{}.
\newblock \showarticletitle{Cartesian genetic programming}.
\newblock In \bibinfo{booktitle}{\emph{Cartesian Genetic Programming}}.
  \bibinfo{publisher}{Springer}, \bibinfo{pages}{17--34}.
\newblock


\bibitem[\protect\citeauthoryear{Olson, La~Cava, Orzechowski, Urbanowicz, and
  Moore}{Olson et~al\mbox{.}}{2017}]%
        {Olson2017PMLB}
\bibfield{author}{\bibinfo{person}{Randal~S. Olson}, \bibinfo{person}{William
  La~Cava}, \bibinfo{person}{Patryk Orzechowski}, \bibinfo{person}{Ryan~J.
  Urbanowicz}, {and} \bibinfo{person}{Jason~H. Moore}.}
  \bibinfo{year}{2017}\natexlab{}.
\newblock \showarticletitle{{PMLB}: a large benchmark suite for machine
  learning evaluation and comparison}.
\newblock \bibinfo{journal}{\emph{BioData Mining}} \bibinfo{volume}{10},
  \bibinfo{number}{1} (\bibinfo{date}{11 Dec} \bibinfo{year}{2017}),
  \bibinfo{pages}{36}.
\newblock


\bibitem[\protect\citeauthoryear{Pedregosa, Varoquaux, Gramfort, Michel,
  Thirion, Grisel, Blondel, Prettenhofer, Weiss, Dubourg, Vanderplas, Passos,
  Cournapeau, Brucher, Perrot, and Duchesnay}{Pedregosa et~al\mbox{.}}{2011}]%
        {scikit-learn}
\bibfield{author}{\bibinfo{person}{F. Pedregosa}, \bibinfo{person}{G.
  Varoquaux}, \bibinfo{person}{A. Gramfort}, \bibinfo{person}{V. Michel},
  \bibinfo{person}{B. Thirion}, \bibinfo{person}{O. Grisel},
  \bibinfo{person}{M. Blondel}, \bibinfo{person}{P. Prettenhofer},
  \bibinfo{person}{R. Weiss}, \bibinfo{person}{V. Dubourg}, \bibinfo{person}{J.
  Vanderplas}, \bibinfo{person}{A. Passos}, \bibinfo{person}{D. Cournapeau},
  \bibinfo{person}{M. Brucher}, \bibinfo{person}{M. Perrot}, {and}
  \bibinfo{person}{E. Duchesnay}.} \bibinfo{year}{2011}\natexlab{}.
\newblock \showarticletitle{Scikit-learn: Machine Learning in {P}ython}.
\newblock \bibinfo{journal}{\emph{Journal of Machine Learning Research}}
  \bibinfo{volume}{12} (\bibinfo{year}{2011}), \bibinfo{pages}{2825--2830}.
\newblock


\bibitem[\protect\citeauthoryear{Pena-Reyes and Sipper}{Pena-Reyes and
  Sipper}{2001}]%
        {Pena2001}
\bibfield{author}{\bibinfo{person}{C.A. Pena-Reyes} {and} \bibinfo{person}{M.
  Sipper}.} \bibinfo{year}{2001}\natexlab{}.
\newblock \showarticletitle{Fuzzy {CoCo}: a cooperative-coevolutionary approach
  to fuzzy modeling}.
\newblock \bibinfo{journal}{\emph{IEEE Transactions on Fuzzy Systems}}
  \bibinfo{volume}{9}, \bibinfo{number}{5} (\bibinfo{year}{2001}),
  \bibinfo{pages}{727--737}.
\newblock


\bibitem[\protect\citeauthoryear{Quade}{Quade}{2020}]%
        {Ohjeah}
\bibfield{author}{\bibinfo{person}{Markus Quade}.}
  \bibinfo{year}{2020}\natexlab{}.
\newblock \bibinfo{title}{cartesian}.
\newblock \bibinfo{howpublished}{\url{https://github.com/Ohjeah/cartesian}}.
\newblock


\bibitem[\protect\citeauthoryear{Sipper and Moore}{Sipper and Moore}{2019}]%
        {sipper2019omnirep}
\bibfield{author}{\bibinfo{person}{Moshe Sipper} {and} \bibinfo{person}{Jason~H
  Moore}.} \bibinfo{year}{2019}\natexlab{}.
\newblock \showarticletitle{{OMNIREP}: originating meaning by coevolving
  encodings and representations}.
\newblock \bibinfo{journal}{\emph{Memetic computing}} \bibinfo{volume}{11},
  \bibinfo{number}{3} (\bibinfo{year}{2019}), \bibinfo{pages}{251--261}.
\newblock


\end{thebibliography}
\bibliographystyle{ACM-Reference-Format}
\end{document}